\pdfoutput=1
\documentclass{article} %
\usepackage{iclr2027_conference,times}
\usepackage{etoolbox}

\makeatletter
\patchcmd{\@maketitle}
  {\begin{tabular}[t]{l}\bf\rule{\z@}{24pt}\@author\end{tabular}}
  {\begin{tabular}[t]{@{}l@{}}\bf\rule{\z@}{24pt}\@author\end{tabular}}
  {}{}
\makeatother

\usepackage{amsmath,amsfonts,bm}

\def\eqref#1{equation~\ref{#1}}

\def\1{\bm{1}}

\DeclareMathAlphabet{\mathsfit}{\encodingdefault}{\sfdefault}{m}{sl}
\SetMathAlphabet{\mathsfit}{bold}{\encodingdefault}{\sfdefault}{bx}{n}

\usepackage{hyperref}
\usepackage{url}
\usepackage{graphicx}
\usepackage{wrapfig}
\usepackage{algorithm,algpseudocode,amsmath}
\usepackage{subcaption}
\usepackage{xcolor}
\usepackage{multirow}
\usepackage{booktabs}
\newcommand\blfootnote[1]{%
  \begingroup
  \renewcommand\thefootnote{}\footnote{#1}%
  \addtocounter{footnote}{-1}%
  \endgroup
}

\title{Triadic Linear Attention: Three-Dimensional Recurrent States for Long-Context Sequence Modeling}

\author{
\makebox[\textwidth][c]{%
\small\bfseries
Oliver Sieberling$^{1}$
\hspace{0.8em}
Bharat Runwal$^{2}$
\hspace{0.8em}
David Jin$^{1}$
\hspace{0.8em}
Ryan Chin$^{1}$
\hspace{0.8em}
Rameswar Panda$^{2}$
\hspace{0.8em}
Yoon Kim$^{1}$
}
\\[7pt]
\makebox[\textwidth][c]{%
\normalfont
$^{1}$Massachusetts Institute of Technology
\hspace{2.5em}
$^{2}$MIT-IBM Computing Research Lab
}
\\[5pt]
\makebox[\textwidth][c]{%
\normalfont\texttt{osieberl@mit.edu}
}
}

\iclrfinalcopy

\begin{document}

\maketitle

\lhead{Preprint}

\begin{abstract}
Recurrent neural networks (RNNs) compress the historical context into a memory state of fixed size, thus allowing for constant-time inference. The memory state size is a crucial factor in their performance, as exemplified by the strong performance and resurgence of linear attention, which extends the vector-valued hidden states of ordinary RNNs to  matrix-valued hidden states. Crucially, linear attention does so in a parameter-efficient way, in particular by using an outer product of the key and value vectors to write to the matrix-valued hidden state. We generalize this construction and propose \emph{triadic linear attention}, which writes the triadic outer product of a key, a second key, and a value, into a  \emph{third-order} (i.e., 3D) tensor state, and reads from it by contracting both key axes with two queries. An $E$-dimensional second key thus yields an $E$-fold increase in state size while adding only two projections. Triadic linear attention is compatible with data-dependent forgetting, the delta rule, and chunkwise-parallel training. Applied to Gated DeltaNet and scalar-gated linear attention, triadic linear attention substantially improves long-context language modeling and recall, outperforming alternatives that enlarge the state.\blfootnote{Code: \url{https://github.com/OliverSieberling/TriadicLinearAttention}.}

\end{abstract}

\section{Introduction} 
Recent works on linear attention have highlighted the importance of \emph{memory state size} in RNNs. Linear attention \citep{katharopoulos2020transformers} extends  vector-valued hidden states in ordinary RNNs to matrix-valued hidden states by using an outer product of two input-dependent vectors (i.e., the key and value vectors) to update the hidden state matrix. The increased  state size has made it possible for modern RNNs to outperform classic RNNs despite making strong structural assumptions on the  recurrence function---needed to parallelize the model across sequence length for efficient training.

State size determines how much of the context a model can recall \citep{arora2024simple}---a model with a fixed-size state cannot even copy sequences beyond a certain length \citep{pmlr-v235-jelassi24a}. Modern linear attention models still struggle to retrieve information from their inputs and degrade on recall-intensive and long-context tasks \citep{arora2024simple, hsieh2024ruler}. However, na\"ively increasing the state size would increase parameter count significantly: adding more heads to the sequence mixer enlarges all its projections, and increasing the value dimension or putting multiple value heads per key still grows the value and output projections.

How can we increase the state size of linear attention in a parameter-efficient way? We  observe that linear attention obtains its matrix-valued state by lifting the vector-valued state of classic RNNs by one order through a \emph{dyadic} outer product of two vectors. This  generalizes to an outer product of $n$ vectors, yielding an $n$-way tensor state. This paper describes \emph{triadic linear attention}, which uses $n=3$ vectors (a key, a second key, and a value) and writes their triadic outer product into a third-order state. It reads from the state by contracting both key axes with two queries. This provides a  parameter-efficient way to increase state size, since an $E$-dimensional second key and query give an \emph{$E$-fold increase} in state size while adding only the two projections that produce them. 

Triadic linear attention is compatible with  three main innovations of modern linear attention variants. First, data-dependent forgetting \citep{peng_2021_random,yang2023gated,Gu2023MambaLS, dao2024transformers} extends naturally by giving each slice along the second-key axis its own forget gate. Second, the delta rule \citep{schlag_linear_2021, yang2024parallelizing,yang2024gated} extends by removing the value stored under both keys jointly before writing the new one. Third, triadic linear attention can be trained efficiently at scale with the chunkwise-parallel form \citep{hua2022transformer, sun2023retentive,yang2024parallelizing}, which we achieve by tiling the state along its value axis, so that no streaming multiprocessor (SM) ever holds the full third-order state of a head.

We apply triadic linear attention to Gated DeltaNet (GDN) and scalar-gated linear attention (sGLA), where we find that  triadic linear attention substantially improves long-context language modeling and recall capabilities (see  Figure~\ref{fig:ladder}). Our approach   outperforms alternative approaches that  also increase the state size through larger heads, larger values, more heads, or multiple value heads per key. 
We further demonstrate that a vanilla linear attention model can be converted to a triadic model post hoc by adapting a pretrained linear attention model to its triadic form during long-context extension. Finally, in a GDN/Transformer hybrid, making the GDN layers triadic lowers perplexity on long books by more than doubling the number of key-value heads does, while using significantly less memory at long context lengths. Together, our results suggest that triadic linear attention is a parameter-efficient way to enlarge state size and improve recurrent neural networks on long-context and recall-intensive tasks.

\section{Triadic Linear Attention}
\subsection{Linear attention as an associative memory}
Consider ordinary single-head  linear attention on a sequence of token representations $\mathbf{x}_1, \dots, \mathbf{x}_T$. Each token produces a query $\mathbf{q}_t$, a key $\mathbf{k}_t$ and a value $\mathbf{v}_t$ through learned linear projections of $\mathbf{x}_t$. At every step $t$, linear attention writes to a recurrent matrix state $\mathbf{S}_t$ through a rank-one update and reads from it by a matrix-vector product:
\begin{equation}
  \mathbf{S}_t = \mathbf{S}_{t-1} + \mathbf{k}_t\,\mathbf{v}_t^{\top},
  \qquad
  \mathbf{o}_t = \mathbf{S}_t^{\top}\mathbf{q}_t
  = \sum_{s\le t}\big(\mathbf{q}_t^{\top}\mathbf{k}_s\big)\,\mathbf{v}_s .
  \label{eq:la}
\end{equation}
This formulation coincides with the fast-weight programmer of \citet{schmidhuber1992learning}, where \emph{slow-weights} emit representations to update a \emph{fast-weight} network \citep{schlag_linear_2021}. Here the fast weights (or equivalently, the hidden states of the RNN) are updated via a \emph{dyadic} {outer product} of two input-dependent vectors. It is also a tensor product representation in the sense of \cite{Smolensky1990Tensor}, where a symbolic structure is embedded by binding each filler to its role through an outer product and storing all bindings in superposition by adding them up. Then, a filler is retrieved from this state by contracting it with a vector that matches the filler's role. In Equation~\ref{eq:la} the value is the filler, the key is the role, $\mathbf{S}_t$ is the superposition, and the readout with the query is the unbinding. A lossless readout requires the query to be orthogonal to the keys of all other stored associations. Otherwise, the retrieval is lossy and every other value interferes in proportion to the overlap of its key and  query. The appeal of this formulation is that storing associations beyond the capacity of the state degrades retrieval gradually, in contrast to other memory systems that evict entire entries.\footnote{Sliding-window attention is a modern instance of such an evicting memory system, since it keeps the last $w$ key-value pairs exactly and removes previous representations entirely.}

The capacity of the state is fully determined by its shape. With $d$-dimensional keys and values, $\mathbf{S}_t$ has $d^2$ entries and can separate at most $d$ mutually orthogonal keys. Beyond this limit, interference is inevitable. Recent advances such as data-dependent forgetting \citep{yang2023gated, dao2024transformers} and the delta rule \citep{widrow_adaptive_1988,schlag_linear_2021, yang2024parallelizing} improve \emph{how} the state is managed, but every linear attention variant remains fundamentally bound by the same $d^2$-entry matrix. To lift this capacity limit, the state itself has to grow.
 
\subsection{Binding to a second key}
The tensor product view suggests a principled way to enlarge the state. As noted by \cite{Smolensky1990Tensor}, a binding itself is a vector (of dimension $d^2$ rather than $d$), and can therefore be bound to a further role. Each such nesting raises the order of the tensor memory by one. One application of this is the third-order tensor product representation of \cite{schlag2018thirdOrder}, which stores a graph and learns its node and relation representations end-to-end. There, each edge is the outer product of its start node, relation and end node, the superposition of these encodes the graph, and unbinding with a start node and a relation retrieves the end node. 

Here, we observe that the third order is useful beyond nested data, as a way to enlarge the capacity of the state. With orthonormal roles, an $n$-th order tensor product representation can store $d^{n-1}$ associations exactly, so binding each value to a second key increases the capacity of linear attention from $d$ to $d^2$ associations. The second key does not correspond to anything in the data and the model can freely choose how to use both keys to store associations. This can be seen as a general way of fast-weight programming a three-dimensional state.

We call the resulting model \emph{triadic linear attention}, since the fast weights (i.e., hidden states) are obtained from a \emph{triadic} outer product of three input-dependent vectors.   Each position produces, besides $\mathbf{q}_t, \mathbf{k}_t, \mathbf{v}_t$ of dimension $d$, a second key $\mathbf{k}'_t$ and second query $\mathbf{q}'_t$ of dimension $E$, also through linear projections of $\mathbf{x}_t$. The state is a third-order tensor $\mathbf{S}_t$ of size $d \times E \times d$, fast-weight programmed through the outer product of three vectors and read by contracting its two key axes with  two queries,
\begin{equation}
  \mathbf{S}_t = \mathbf{S}_{t-1} + \mathbf{k}_t\otimes\mathbf{k}'_t\otimes\mathbf{v}_t,
  \qquad
  \mathbf{o}_t = \mathbf{S}_t\times_1\mathbf{q}_t\times_2\mathbf{q}'_t
  = \sum_{s\le t}\big(\mathbf{q}_t^{\top}\mathbf{k}_s\big)\big(\mathbf{q}'^{\top}_t\mathbf{k}'_s\big)\,\mathbf{v}_s ,
  \label{eq:la3}
\end{equation}
where $\times_n$ denotes contraction along axis $n$. In the language of the previous subsection, the value is bound to two roles at once and unbound with two queries. For $E=1$ with $\mathbf{k}'_t=\mathbf{q}'_t=1$, Equation~\ref{eq:la3} reduces to Equation~\ref{eq:la}, so ordinary linear attention is a special case.

The state now has $d^2 E$ entries instead of $d^2$, which is cubic in $d$ for $E=d$. Meanwhile, the parameters grow only by the two projections needed to produce $\mathbf{k}'_t$ and $\mathbf{q}'_t$, and therefore the ratio of state size to parameter count increases asymptotically. In practice, we keep $E$ small (e.g., $E=8$), which makes the parameter overhead negligible while multiplying the capacity of the state by $E$.

\subsection{Capacity in isolation}

Before adding further machinery, we verify that the construction of Equation~\ref{eq:la3} actually expands the capacity. To this end, we use multi-query associative recall (MQAR; \citealp{arora2023zoology}),\begin{wrapfigure}{r}{0.48\textwidth}
    \centering
    \vspace{-1mm}
    \includegraphics[width=0.48\textwidth]{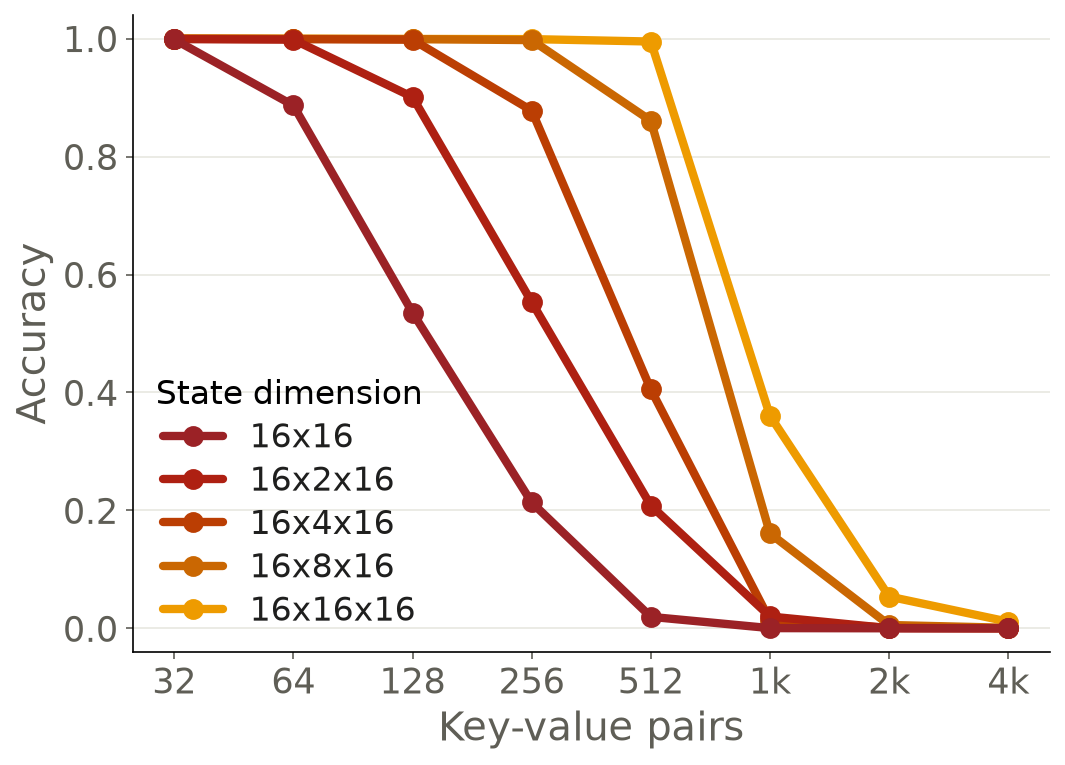}
    \vspace{-4mm}
    \caption{MQAR with plain linear attention (state 16$\times$16) and triadic linear attention with $E\! \in \! \{2,4,8,16\}$ (state 16$\times$E$\times$16).}
    \label{fig:mqar}
\end{wrapfigure}where each example contains $N$ key-value pairs followed by the same $N$ keys in random order, and the model must predict the value corresponding to each key. We keep the model deliberately minimal, with two layers of four heads with $d=16$ and a sequence mixer that is Equation~\ref{eq:la3}, without convolutions, forgetting, non-linearities, or the delta rule. We compare plain linear attention ($E{=}1$) to triadic linear attention with $E \in \{2,4,8,16\}$ across $N$ from 32 to 4096. The detailed setup can be found in Appendix~\ref{app:mqarsetup}.
Figure~\ref{fig:mqar} shows that every doubling of $E$ shifts the accuracy curve to the right by roughly a doubling of $N$, e.g., for $E=16$ the model stores roughly 16 times as many key-value associations as ordinary linear attention. Notably, this 16-fold increase in state capacity comes with only a 1.08-fold increase in non-embedding parameters.

\subsection{Forgetting and the delta rule}
\label{subsec:method:forgetting-delat}
Modern linear attention variants manage the state with data-dependent forgetting \citep{yang2023gated, dao2024transformers} and the delta rule \citep{widrow_adaptive_1988, schlag_linear_2021, yang2024parallelizing}, which removes (part of) the previously stored association before associating a key with a new value. We extend both ideas to the three-dimensional state.

\paragraph{Forgetting.}
Linear attention with scalar decay computes a gate $\alpha_t \in [0,1]$ and multiplies the recurrent state with it before applying the outer product update. For the three-dimensional state we vary the forgetting along the second-key axis by giving each slice $\mathbf{S}_t[:,e,:]$ its own decay gate $\alpha_{t,e}$, yielding the recurrence
\begin{equation}
  \mathbf{S}_t = \mathbf{S}_{t-1}\times_2\operatorname{diag}(\boldsymbol{\alpha}_t) + \mathbf{k}_t\otimes\mathbf{k}'_t\otimes\mathbf{v}_t .
  \label{eq:forget3}
\end{equation}
Here, within a slice the decay is scalar, and across slices it is channelwise in the sense of gated linear attention \citep{yang2023gated}. This comes with little parameter overhead, even when parametrized with a dense linear layer, since the second-key axis is small.

\paragraph{Delta rule.}
Given a new key-value pair, the delta rule first removes what the state currently stores for the key, before writing the new association. For triadic linear attention, the currently stored association is the two-key readout $\mathbf{S}_{t-1} \times_1 \mathbf{k}_t \times_2 \mathbf{k}'_t$. Removing exactly that value and interpolating toward the new one gives  
\begin{equation}
  \mathbf{S}_t = \mathbf{S}_{t-1}
  + \beta_t\,\mathbf{k}_t\otimes\mathbf{k}'_t\otimes\big(\mathbf{v}_t - \mathbf{S}_{t-1}\times_1\mathbf{k}_t\times_2\mathbf{k}'_t\big),
  \label{eq:gdn3}
\end{equation}
with a data-dependent write strength $\beta_t \in (0,1)$. This can be combined with forgetting by decaying $\mathbf{S}_{t-1}$ before the erase, as in Equation~\ref{eq:forget3}.

Equation~\ref{eq:gdn3} applies the delta rule jointly to the pair of keys. By flattening the outer product of both keys into a joint key $\boldsymbol{\kappa}_t = \mathbf{k}_t \otimes \mathbf{k}'_t$ of dimension $d \cdot E$ and interpreting the state as a $d\cdot E \times d$ matrix, we can rewrite the update as
\begin{equation}
  \mathbf{S}_t = \big(\mathbf{I} - \beta_t\,\boldsymbol{\kappa}_t\boldsymbol{\kappa}_t^{\top}\big)\mathbf{D}_t\,\mathbf{S}_{t-1} + \beta_t\,\boldsymbol{\kappa}_t\mathbf{v}_t^{\top},
  \qquad
  \mathbf{o}_t = \mathbf{S}_t^{\top}\big(\mathbf{q}_t\otimes\mathbf{q}'_t\big),
  \label{eq:gdn3flat}
\end{equation}
where $\mathbf{D}_t$ is a diagonal decay matrix. This corresponds to Gated DeltaNet with a key dimension of $d \cdot E$, read with the $d \cdot E$-dimensional joint query $\mathbf{q}_t \otimes \mathbf{q}'_t$, and with a per-slice scalar decay.

\subsection{Efficient implementation}
\label{subsec:efficient}
Linear attention trains efficiently with the \emph{chunkwise-parallel form} \citep{hua2022transformer, sun2023retentive, yang2024parallelizing}. 
The sequence is split into chunks of $C$ positions, and $\square^r$ denotes the quantity $\square$ at position $r$ of the current chunk. 
With the chunk's queries, keys and values stacked into the rows of $\mathbf{Q}, \mathbf{K}, \mathbf{V} \in \mathbb{R}^{C \times d}$ and $\mathbf{S}_{[i]}$ the state at the start of the chunk,  linear attention computes
\begin{equation}
  \mathbf{O} = \mathbf{Q}\,\mathbf{S}_{[i]}
  + \operatorname{tril}\!\big(\mathbf{Q}\mathbf{K}^{\top}\big)\,\mathbf{V},
  \qquad
  \mathbf{S}_{[i+1]} = \mathbf{S}_{[i]} + \mathbf{K}^{\top}\mathbf{V},
  \label{eq:chunk}
\end{equation}
so all positions of a chunk interact in parallel through masked attention, and
only the state at chunk boundaries is carried by a recurrence. 

Triadic linear attention is linear attention with the $d \cdot E$-dimensional joint key $\mathbf{k}_t \otimes \mathbf{k}'_t$, so Equation~\ref{eq:chunk} applies in principle. 
Treating it this way, however, would cost $E$ times more in every inner product of the masked attention, and $E$ times more in the state, which no longer fits on chip. 
The Kronecker structure of the joint key removes the first cost, and tiling the state addresses the on-chip memory constraint.
We describe the chunkwise parallel form for triadic linear attention without forgetting or the delta rule here and defer the complete chunkwise parallel forms to Appendix~\ref{app:kernels}.

\paragraph{Separating the joint key.}
Inner products of joint queries and keys factorize,
$(\mathbf{q}^r \otimes \mathbf{q}'^r)^{\top}(\mathbf{k}^s \otimes \mathbf{k}'^s)
= (\mathbf{q}^{r\top}\mathbf{k}^s)(\mathbf{q}'^{r\top}\mathbf{k}'^s)$, so with
$\mathbf{Q}', \mathbf{K}' \in \mathbb{R}^{C \times E}$ the masked attention of
the chunk is
\begin{equation}
  \big(\mathbf{Q}\mathbf{K}^{\top}\big) \odot \mathbf{R}',
  \qquad
  \mathbf{R}' = \operatorname{tril}\!\big(\mathbf{Q}'\mathbf{K}'^{\top}\big).
  \label{eq:separate}
\end{equation}
The interactions within a chunk reduce to ordinary linear attention with $d$-dimensional keys, whose causal mask is replaced by the $C \times C$ matrix $\mathbf{R}'$. 
The masked attention therefore works with $C \times C$ matrices, as in ordinary linear attention, and forming the matrix costs $C^2 (d + E)$ operations instead of the $C^2 (d \cdot E)$ of the joint keys. 
The only extra work over ordinary linear attention is the $C^2 E$ of computing $\mathbf{R}'$.

\paragraph{Tiling the state.}
The $E$-fold cost is thus confined to the state. Writing the
$d \times E \times d$ state as slices $\mathbf{S}_{[i],e} \in
\mathbb{R}^{d \times d}$ and $\mathbf{q}'_e, \mathbf{k}'_e \in \mathbb{R}^{C}$
for the columns of $\mathbf{Q}', \mathbf{K}'$, the chunk reads and writes each
slice with the shared queries and keys,
\begin{equation}
  \mathbf{O} = \sum_{e=1}^{E} \operatorname{diag}(\mathbf{q}'_e)\,\mathbf{Q}\,\mathbf{S}_{[i],e}
  + \big(\mathbf{Q}\mathbf{K}^{\top} \odot \mathbf{R}'\big)\,\mathbf{V},
  \qquad
  \mathbf{S}_{[i+1],e} = \mathbf{S}_{[i],e}
  + \mathbf{K}^{\top}\operatorname{diag}(\mathbf{k}'_e)\,\mathbf{V}.
  \label{eq:slices}
\end{equation}
The slices interact only through the sum over $e$, and different columns of the value axis never interact. 
Our kernels therefore split the state of each head along the value axis into blocks of 32 columns, one per thread block, and each thread block keeps all $E$ slices of its columns in registers for the entire sequence, so the full state is never held in one place. At $E = 8$, the state of one head occupies 512\,KiB in FP32, twice the register file of a Hopper streaming multiprocessor, whereas one block of it occupies 128\,KiB.
Section~\ref{subsec:efficiency} measures the resulting cost, and Appendix~\ref{app:kernels} describes the kernels.

\begin{figure}[t]
  \centering
\vspace{-0mm}
  \begin{subfigure}[t]{0.42\linewidth}
    \centering
    \includegraphics[width=\linewidth]{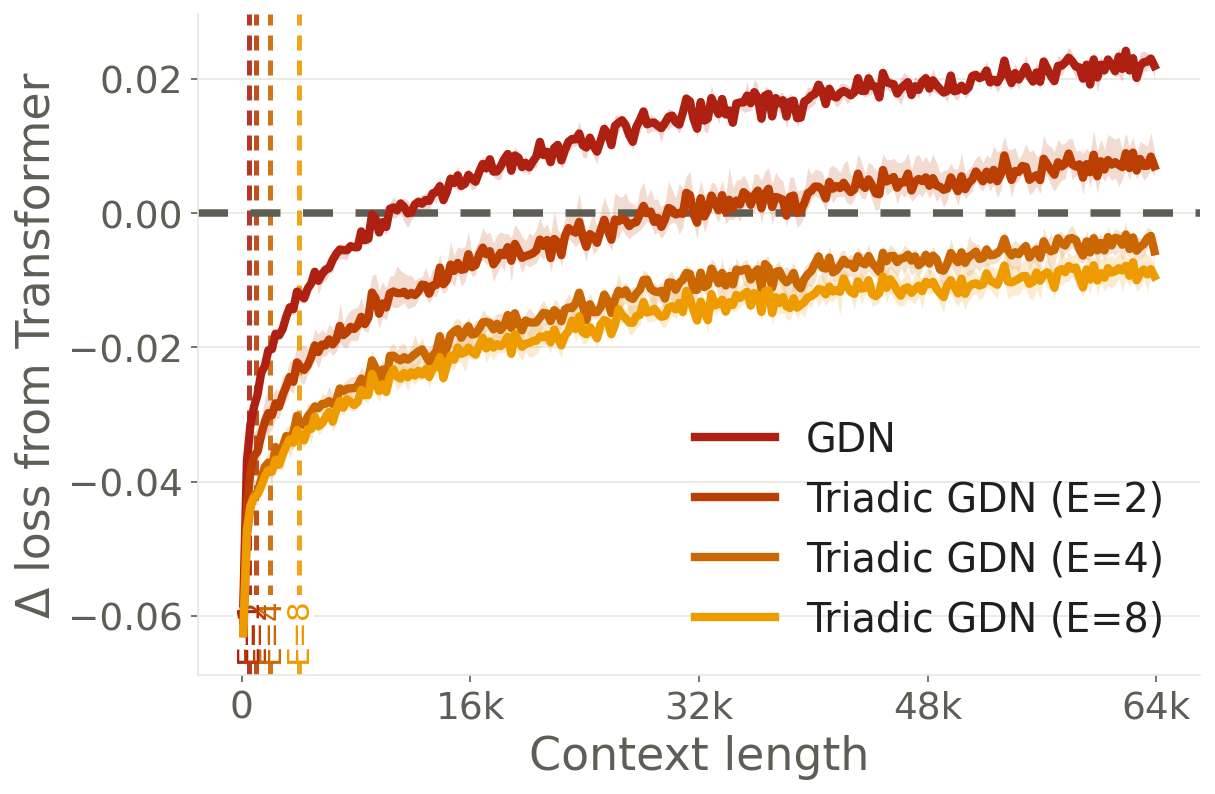}
    \label{fig:ladder-ctx-400M}
  \end{subfigure}
  \hfill
  \begin{subfigure}[t]{0.56\linewidth}
    \centering
    \includegraphics[width=\linewidth]{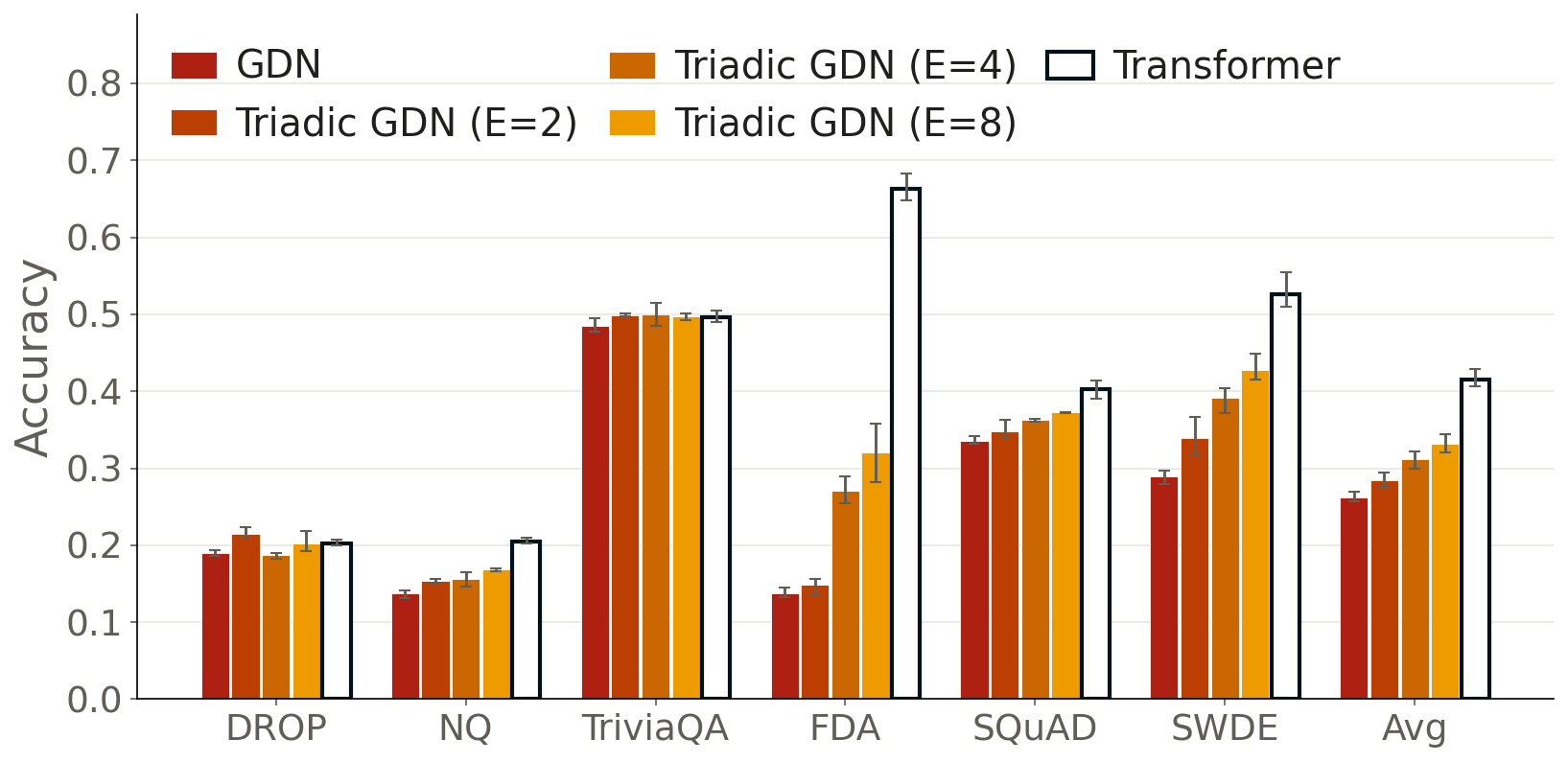}
    \label{fig:ladder-recall-400M}
  \end{subfigure}

  \vspace{0.2em}

  \begin{subfigure}[t]{0.42\linewidth}
    \centering
    \includegraphics[width=\linewidth]{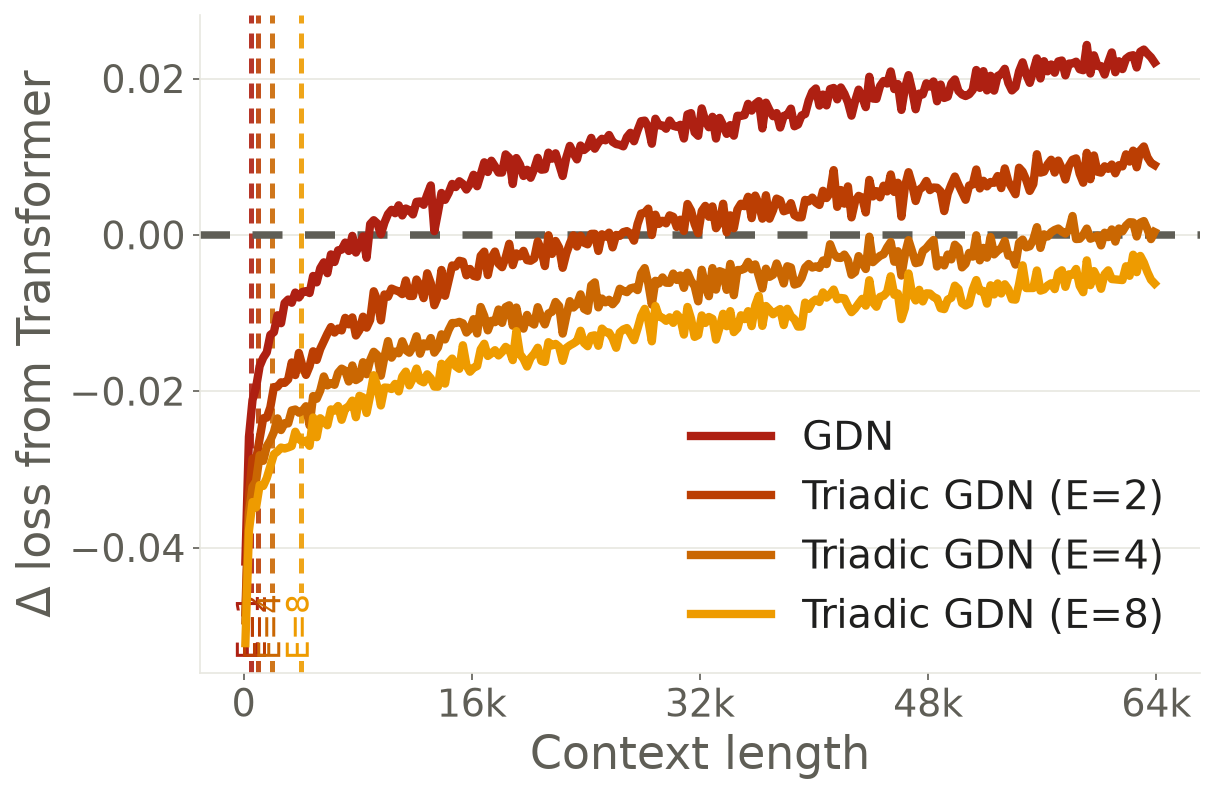}
    \label{fig:ladder-ctx-1p3B}
  \end{subfigure}
  \hfill
  \begin{subfigure}[t]{0.56\linewidth}
    \centering
    \includegraphics[width=\linewidth]{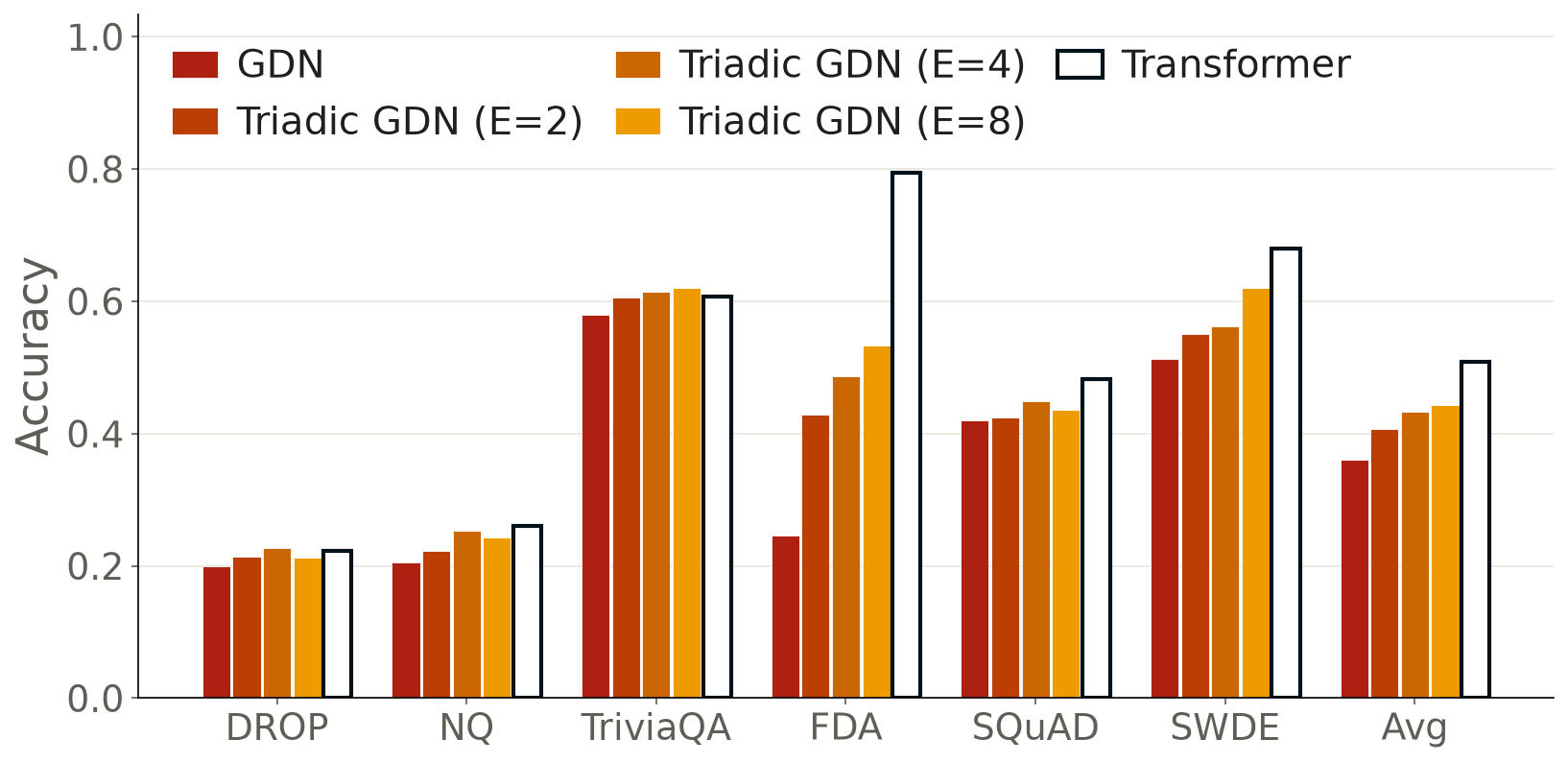}
    \label{fig:ladder-recall-1p3B}
  \end{subfigure}
\vspace{-1mm}
  \caption{
    Triadic Gated DeltaNet with a growing second key dimension $E$ at
    400M parameters (top) and 1.3B parameters (bottom).
    Left: Loss difference to the Transformer on PG19 by context position,
    with dashed lines marking where the key-value cache of the Transformer
    exceeds the state of each model.
    Right: Accuracy on six recall-intensive tasks.
  }
  \label{fig:ladder}
  \vspace{-1mm}
\end{figure}

\section{Empirical Study}
\label{sec:empiricalstudy}

\subsection{Experimental Setup}
\paragraph{Models.}
We train models at 400M parameters (24 layers, $d_{\text{model}} = 1024$, 8 heads) and 1.3B parameters (24 layers, $d_{\text{model}} = 2048$, 16 heads), with head dimension $d=128$ throughout. We apply triadic linear attention to two base sequence mixers, Gated DeltaNet (GDN; \citealp[]{yang2024gated}) and scalar-gated linear attention (sGLA), which we define as GDN without the erase term of the delta rule.\footnote{sGLA thereby belongs to the same family as other linear attention models with a data-dependent scalar decay, such as gated RFA \citep{peng_2021_random}, Mamba-2 \citep{dao2024transformers} and  mLSTM \citep{beckxlstm}. However, Mamba-2 and xLSTM have other components which we do not make use of, and hence we use sGLA to describe linear attention with scalar data-dependent gates.} The triadic variants keep the base mixer and add a second key and query, which are obtained through a linear projection followed by a short convolution and a softplus activation. Additionally, each slice obtains a separate scalar forget gate, as described in Section~\ref{subsec:method:forgetting-delat}. The detailed architecture is described in Appendix~\ref{app:lmsetup}. For $E=1$, Triadic GDN reduces exactly to Gated DeltaNet, while $E=8$ adds only 1.2\% more parameters. We train $E \in \{1,2,4,8\}$ for both variants at 400M and for Triadic GDN at 1.3B.

\paragraph{Training.}
We pretrain each model on 50 tokens per parameter, i.e., 2.5$\times$ Chinchilla-optimal \citep{hoffmann2022training}, which corresponds to 20B tokens for the 400M parameter models and 65B tokens for the 1.3B parameter models. We train on Fineweb-Edu \citep{lozhkov2024fineweb-edu} with context length 4k, then long-context extend all models to 64k on 5 tokens per parameter using a mixture of Fineweb-Edu, PG19 books data \citep{rae2019compressive} and scientific PDFs \citep{olmo2026olmo3}. We use the AdamW optimizer \citep{loshchilov2019decoupledweightdecayregularization} with peak learning rate $3 \times 10^{-4}$ for pretraining and peak learning rate $10^{-4}$ for long context extension, with weight decay $0.1$ throughout. The learning rate is warmed up linearly and then decays with a cosine schedule to 10\% of its peak. For both training stages, we use an effective batch size of about 0.5M tokens at 400M parameters and about 1M tokens at 1.3B parameters.

\paragraph{Baselines.}
In addition to ordinary sGLA/GDN baselines,  we compare against established methods for increasing the state size, compared at matched state size. Specifically, we train four variants of each model with $2\times$ and $4\times$ its state through a larger head dimension with fewer heads (\emph{larger heads} in Table 1), a larger value dimension (\emph{wider values}), several value heads per key head (\emph{grouped values}), or more heads (\emph{more heads}). A larger head dimension leaves the total width of projections unchanged, while the other variants enlarge the linear projections of the sequence mixer significantly. We account for this by reducing the MLP width to match the parameter count. Adding heads is only feasible at $2\times$, since at $4\times$ the projections of the sequence mixer alone exceed the parameter budget. For reference, we also train Transformers with RoPE \citep{su2021roformer}, QK-norm \citep{pmlr-v202-dehghani23a} and grouped-query attention \citep{ainslie2023gqa} with eight query heads per key-value head. We use a RoPE base frequency of 10k during pretraining and raise it to 2M for long-context extension.

\paragraph{Evaluation.}
We evaluate all models after long-context extension. We report perplexity on held-out PG19 books \citep{rae2019compressive} by context position and perplexity on Wikitext \citep{merity2016pointer}. To measure in-context recall, we use the benchmark suite from \citet{arora2024simple}, and additionally report the average over ten zero-shot tasks. More details are given in Appendix~\ref{app:lmsetup}.
\begin{table}[t]
\centering
\scriptsize
\setlength{\tabcolsep}{3.2pt}
\renewcommand{\arraystretch}{1.08}
\caption{Different ways of enlarging the state size of Gated DeltaNet (GDN) and scalar-gated linear attention (sGLA) by $2\times$ and $4\times$ at matched parameter count. State in MB (assuming 2 bytes per entry), and parameters in millions. Results are averaged over three training seeds.}
\label{tab:unified_400m}
\resizebox{\linewidth}{!}{
\begin{tabular}{l|c|c|cc|ccc|c}
\toprule
\textbf{Model}
& \textbf{State}
& \textbf{Params}
& \textbf{0-shot $\uparrow$}
& \textbf{Wiki. $\downarrow$}
& \textbf{PG19 $\leq$4k $\downarrow$}
& \textbf{PG19 4k--16k $\downarrow$}
& \textbf{PG19 16k--64k $\downarrow$}
& \textbf{Recall $\uparrow$} \\
\midrule

Transformer
& 12.6/1k tok
& 377.0
& 52.6 & 11.11
& 15.39 & 14.41 & 13.92
& 41.6 \\

\midrule

GDN (base)
& 6.3
& 380.9
& 53.1 & 11.25
& 15.03 & 14.38 & 14.15
& 26.2 \\

\addlinespace[2pt]

\quad Larger heads ($d {=} 256$)
& 12.6
& 380.7
& 52.6 & 11.22
& 15.10 & 14.42 & 14.17
& 26.6 \\

\quad Grouped values (2 per key)
& 12.6
& 378.2
& 52.9 & 11.18
& 15.05 & 14.37 & 14.12
& 27.1 \\

\quad Wider values ($d_v {=} 256$)
& 12.6
& 377.9
& 53.2 & 11.20
& 15.07 & 14.39 & 14.14
& 27.4 \\

\quad More heads (16)
& 12.6
& 381.6
& 53.3 & 11.32
& 15.25 & 14.54 & 14.28
& 27.5 \\

\quad Triadic ($E {=} 2$)
& 12.6
& 381.9
& \textbf{53.4} & \textbf{11.08}
& \textbf{14.90} & \textbf{14.21} & \textbf{13.95}
& \textbf{28.4} \\

\addlinespace[3pt]

\quad Larger heads ($d {=} 512$)
& 25.2
& 380.6
& 53.0 & 11.25
& 15.21 & 14.50 & 14.24
& 28.5 \\

\quad Grouped values (4 per key)
& 25.2
& 382.4
& 52.7 & 11.35
& 15.35 & 14.63 & 14.36
& 27.3 \\

\quad Wider values ($d_v {=} 512$)
& 25.2
& 381.3
& 52.5 & 11.41
& 15.41 & 14.70 & 14.44
& 27.5 \\

\quad Triadic ($E {=} 4$)
& 25.2
& 383.0
& \textbf{53.6} & \textbf{10.93}
& \textbf{14.80} & \textbf{14.08} & \textbf{13.79}
& \textbf{31.1} \\

\midrule

sGLA (base)
& 6.3
& 380.9
& 53.5 & 11.41
& 15.19 & 14.55 & 14.32
& 26.4 \\

\addlinespace[2pt]

\quad Larger heads ($d {=} 256$)
& 12.6
& 380.7
& 52.4 & 11.38
& 15.22 & 14.57 & 14.33
& 28.2 \\

\quad Grouped values (2 per key)
& 12.6
& 378.2
& 53.0 & 11.34
& 15.19 & 14.53 & 14.28
& \textbf{28.5} \\

\quad Wider values ($d_v {=} 256$)
& 12.6
& 377.9
& 53.5 & 11.35
& 15.19 & 14.53 & 14.28
& 27.7 \\

\quad More heads (16)
& 12.6
& 381.6
& 52.9 & 11.42
& 15.36 & 14.67 & 14.42
& 27.8 \\

\quad Triadic ($E {=} 2$)
& 12.6
& 381.9
& \textbf{53.5} & \textbf{11.19}
& \textbf{15.00} & \textbf{14.32} & \textbf{14.07}
& 28.4 \\

\addlinespace[3pt]

\quad Larger heads ($d {=} 512$)
& 25.2
& 380.6
& 52.2 & 11.52
& 15.53 & 14.85 & 14.59
& 28.7 \\

\quad Grouped values (4 per key)
& 25.2
& 382.4
& 52.8 & 11.47
& 15.50 & 14.79 & 14.52
& 28.3 \\

\quad Wider values ($d_v {=} 512$)
& 25.2
& 381.3
& 52.4 & 11.51
& 15.52 & 14.82 & 14.55
& 29.0 \\

\quad Triadic ($E {=} 4$)
& 25.2
& 383.0
& \textbf{53.4} & \textbf{11.09}
& \textbf{14.94} & \textbf{14.24} & \textbf{13.97}
& \textbf{30.7} \\

\bottomrule
\end{tabular}
}
\vspace{-2mm}
\end{table}

\subsection{Main Results}
\begin{figure*}[t]
  \centering
  \includegraphics[width=\textwidth]{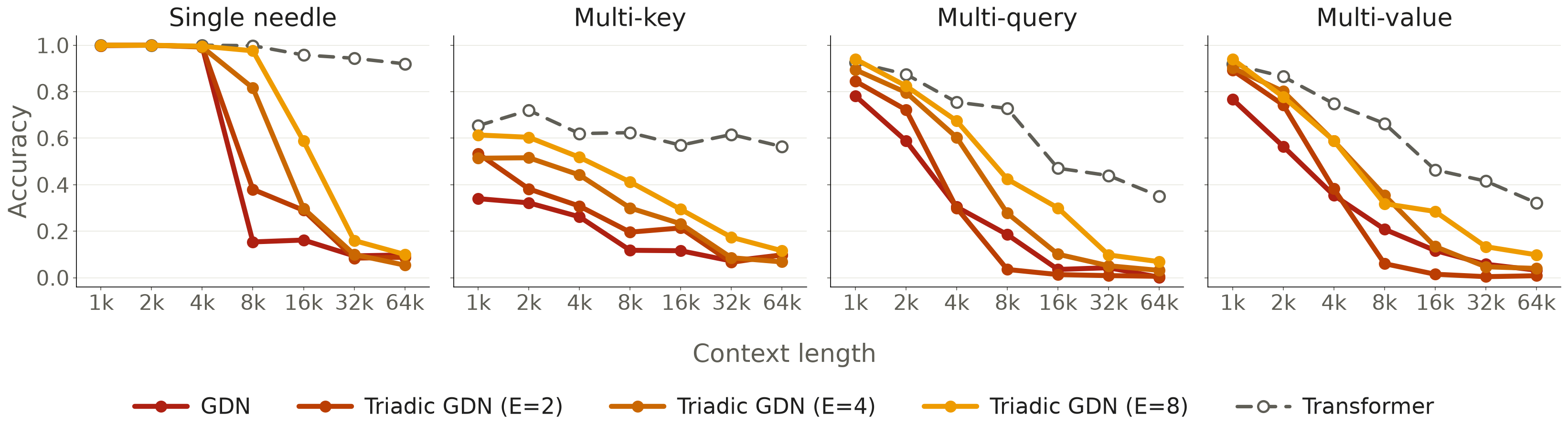}
    \vspace{-3mm}
  \caption{RULER needle-in-a-haystack accuracy by context length at 1.3B parameters.}
  \label{fig:niah_13b}
  \vspace{-0mm}
\end{figure*}

\paragraph{Scaling the state.}
Figure~\ref{fig:ladder} shows how the state size of triadic linear attention affects language modeling and in-context recall at the 400M and 1.3B parameter scales. Gated DeltaNet predicts early tokens better than the Transformer but falls behind at a context length of around 10k tokens. With $E=2$, triadic Gated DeltaNet pushes this crossover point considerably further out, and with $E=8$, it predicts the next token better than the Transformer even at 64k context. The dashed vertical lines indicate the context lengths at which the key-value cache of the Transformer exceeds the state size of the recurrent model, so beyond this point, triadic Gated DeltaNet predicts the next token better with a smaller state. Overall, we observe language modeling to improve as the state size grows, but with diminishing returns. On recall-intensive tasks, triadic Gated DeltaNet substantially improves performance at both scales, with particularly large gains on FDA and SWDE, which require copying information from a long document. Figure~\ref{fig:niah_13b} shows the same trend on the needle-in-a-haystack tasks of RULER \citep{hsieh2024ruler}, where a larger state keeps retrieval accurate up to longer contexts. Appendix~\ref{app:analysis} shows that the larger state is indeed used for distant context, as Triadic GDN with $E=8$ benefits more than GDN from additional context when predicting the same final 4k tokens.
\vspace{-0mm}

\paragraph{State-matched comparison.}
Table~\ref{tab:unified_400m} compares triadic linear attention to alternative methods for increasing the state size. At both $2\times$ and $4\times$, triadic Gated DeltaNet achieves consistently lower perplexity on WikiText and on every PG19 range, while scoring the highest average on the recall evaluation suite. The alternative methods barely improve upon the vanilla Gated DeltaNet at $2\times$, and at $4\times$ they even degrade on every PG19 range. There, two of the alternatives (larger values, more value heads) require so many additional parameters that the MLP width has to be reduced from 2816 to 640 to stay within the parameter budget. Triadic linear attention instead adds only two small projections, which keeps the parameter overhead low and preserves the full MLP. The comparison for sGLA is similar, with triadic sGLA achieving the lowest perplexity at both state sizes.

\subsection{Upcycling to Triadic Linear Attention via Continued Pretraining }
\begin{table}[t]
\centering

\scriptsize
\setlength{\tabcolsep}{3.2pt}
\renewcommand{\arraystretch}{1.08}
\caption{Upcycling a pretrained model from $E=1$ to $E=8$ during long-context extension, compared with the corresponding base model and with $E=8$ pretrained from scratch. Results are averaged over three seeds at 400M and reported for a single seed at 1.3B.}

\label{tab:upcycling}
\resizebox{\linewidth}{!}{
\begin{tabular}{l|c|c|cc|ccc|c}
\toprule
\textbf{Model}
& \textbf{State}
& \textbf{Params}
& \textbf{0-shot $\uparrow$}
& \textbf{Wiki. $\downarrow$}
& \textbf{PG19 $\leq$4k $\downarrow$}
& \textbf{PG19 4k--16k $\downarrow$}
& \textbf{PG19 16k--64k $\downarrow$}
& \textbf{Recall $\uparrow$} \\
\midrule

GDN (base)
& 6.3
& 380.9
& 53.1 & 11.25
& 15.03 & 14.38 & 14.15
& 26.2 \\

\quad Triadic ($E{=}8$), from scratch
& 50.3
& 385.4
& \textbf{53.4} & \textbf{10.87}
& \textbf{14.79} & \textbf{14.04} & \textbf{13.74}
& \textbf{33.1} \\

\quad Triadic ($E{=}8$), upcycled
& 50.3
& 385.4
& 53.2 & 11.04
& 14.91 & 14.18 & 13.86
& 29.7 \\

\midrule

sGLA (base)
& 6.3
& 380.9
& \textbf{53.5} & 11.41
& 15.19 & 14.55 & 14.32
& 26.4 \\

\quad Triadic ($E{=}8$), from scratch
& 50.3
& 385.4
& 53.4 & \textbf{11.03}
& \textbf{14.96} & \textbf{14.23} & \textbf{13.95}
& \textbf{32.1} \\

\quad Triadic ($E{=}8$), upcycled
& 50.3
& 385.4
& 53.3 & 11.16
& 15.04 & 14.32 & 14.03
& 30.0 \\

\midrule

GDN (base)
& 12.6
& 1360.2
& 59.6 & 8.08
& 10.91 & 10.38 & 10.20
& 36.1 \\

\quad Triadic ($E{=}8$), from scratch
& 100.7
& 1378.3
& \textbf{60.9} & \textbf{7.87}
& \textbf{10.74} & \textbf{10.16} & \textbf{9.94}
& \textbf{44.4} \\

\quad Triadic ($E{=}8$), upcycled
& 100.7
& 1378.3
& 59.8 & 7.96
& 10.82 & 10.24 & 10.00
& 42.0 \\

\midrule

sGLA (base)
& 12.6
& 1359.4
& 60.3 & 8.26
& 11.10 & 10.61 & 10.45
& 37.0 \\

\quad Triadic ($E{=}8$), from scratch
& 100.7
& 1377.5
& \textbf{60.4} & \textbf{7.94}
& \textbf{10.82} & \textbf{10.25} & \textbf{10.03}
& \textbf{45.1} \\

\quad Triadic ($E{=}8$), upcycled
& 100.7
& 1377.5
& 60.4 & 8.12
& 10.97 & 10.40 & 10.19
& 42.2 \\

\bottomrule
\end{tabular}
}

\end{table}

We also test whether an ordinary linear attention model can be made triadic after pretraining. To this end, we expand a pretrained model from $E=1$ to $E=8$ by copying its forget gates to every slice and initializing the second key and query projections as well as their convolutions from scratch. Then, we perform the long-context extension on the resulting triadic model. Table~\ref{tab:upcycling} shows that the upcycled models consistently improve over their base models for both sequence mixers and at both scales. They recover about half to three quarters of the gain that a triadic model pretrained from scratch achieves over its dyadic counterpart. Enlarging the state when training on longer context lengths is reminiscent of how the key-value cache of softmax attention grows with the context. Since short-context data requires little state but makes up most of the pretraining data, a linear attention model could potentially be pretrained with a small state that is enlarged in later training stages with higher sequence lengths.

\subsection{Hybrid Models}
\begin{table}[t]
\centering

\scriptsize
\setlength{\tabcolsep}{3.2pt}
\renewcommand{\arraystretch}{1.08}
\caption{Enlarging either the key-value cache (GQA-4) or the linear-attention state (Triadic, $E = 4$) in a 3:1 GDN/GQA-8 hybrid at 400M parameters. State is reported in MB at 4k and 64k tokens. NIAH is averaged over eight RULER tasks. Results are averaged over three seeds.}

\label{tab:hybrid_400m}
\resizebox{\linewidth}{!}{
\begin{tabular}{l|cc|cc|cc|c|c}
\toprule
\textbf{Model}
& \textbf{State @ 4k $\downarrow$}
& \textbf{State @ 64k $\downarrow$}
& \textbf{0-shot $\uparrow$}
& \textbf{Wiki. $\downarrow$}
& \textbf{PG19 $\leq$4k $\downarrow$}
& \textbf{PG19 4k--64k $\downarrow$}
& \textbf{Recall $\uparrow$}
& \textbf{NIAH Avg. $\uparrow$} \\
\midrule

Transformer
& 50 & 805
& 52.6 & 11.11
& 15.39 & 14.02
& 41.6
& 48.2 \\

\midrule

GDN
& 6 & 6
& 53.1 & 11.25
& 15.03 & 14.19
& 26.2
& 21.7 \\

\midrule

3:1 GDN/GQA-8 (base)
& 17 & 206
& 53.0 & 10.69
& 14.86 & 13.50
& 43.7
& 51.5 \\

\quad Larger KV cache (GQA-4)
& 30 & 407
& 53.5 & 10.67
& 14.83 & 13.47
& \textbf{45.3}
& 53.0 \\

\quad Triadic state ($E {=} 4$)
& 31 & 220
& \textbf{53.8} & \textbf{10.63}
& \textbf{14.76} & \textbf{13.39}
& 44.5
& \textbf{56.3} \\

\bottomrule
\end{tabular}
}

\end{table}

In practice, linear attention variants are often deployed interleaved with softmax attention blocks \citep{kimiteam2025kimilinearexpressiveefficient, yang2025qwen3technicalreport}. This raises a natural question: Is it more beneficial to enlarge the state of linear attention blocks or the key-value cache of the softmax attention blocks? To test this, we train a 3:1 GDN/GQA-8 hybrid with NoPE \citep{3666122.3667204}, where GQA-$g$ denotes grouped-query attention with $g$ query heads per key-value head, and compare it to the same model with twice as large key-value cache (3:1 GDN/GQA-4) and to the same model with four times larger linear attention states (3:1 Triadic GDN/GQA-8 ($E{=}4$)). Table~\ref{tab:hybrid_400m} shows that the triadic hybrid achieves the lowest perplexity at every context range as well as the best zero-shot and NIAH averages, and trails 3:1 GDN/GQA-4 only on recall (within seed spread). At the same time, 3:1 GDN/GQA-4 requires more memory than the triadic hybrid beyond about 4.6k tokens and almost twice as much at 64k tokens.

\subsection{Training Efficiency}
\label{subsec:efficiency}
\begin{wrapfigure}{r}{0.52\textwidth}
    \centering
    \vspace{-13mm}
    \includegraphics[width=0.52\textwidth]{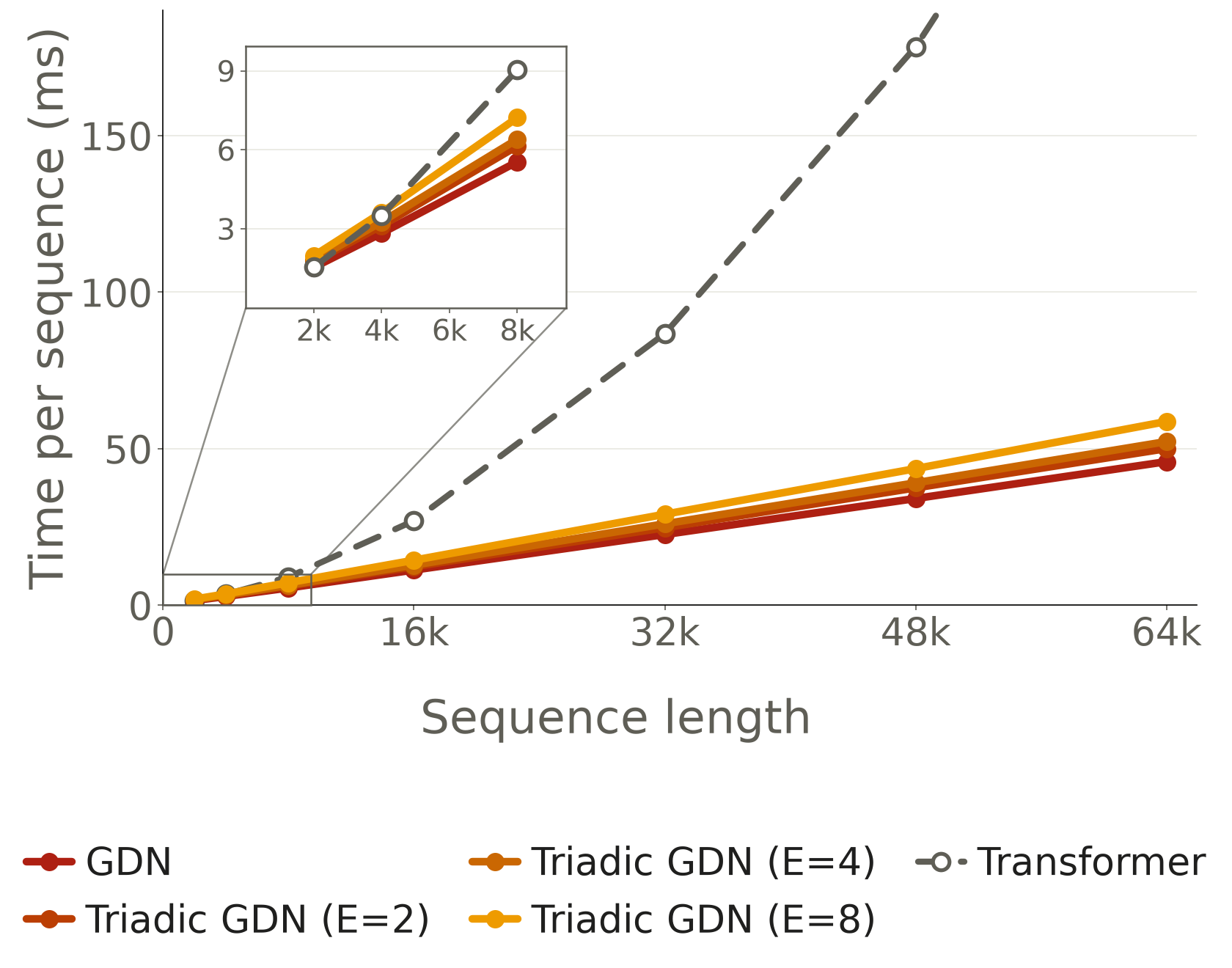}
    \vspace{-6mm}
      \caption{Forward and backward time of one block of the 1.3B models per sequence, at batch size 4 on H100 GPUs. Triadic GDN with $E=8$ adds moderate overhead over GDN and is faster than the Transformer beyond 4k tokens.}
    \label{fig:seqlen}
    \vspace{-2mm}
\end{wrapfigure}

Figure~\ref{fig:seqlen} compares the forward and backward time of one block per sequence of the 1.3B models from 2k to 64k tokens at batch size 4, measured on a single H100.  
Triadic GDN uses our CuTe kernels (Section~\ref{subsec:efficient} and Appendix~\ref{app:kernels}), GDN uses the TileLang kernels of FlashQLA \citep{flashqla2025}, and the Transformer uses FlexAttention \citep{dong2024flexattentionprogrammingmodel}.
GDN matches the Transformer at 2k tokens, and Triadic GDN overtakes it at 4k for $E = 2$ and $E=4$ and at 8k for $E = 8$.
With $E = 8$, it is 3\% slower than the Transformer at 4k and faster at every longer context, reaching 5.1 times faster at 64k. In general, the overhead of Triadic GDN over vanilla Gated DeltaNet is 28\%--30\% for $E=8$, 14\%--15\% for $E=4$, and 9\%--11\% for $E=2$.

\subsection{Ablations}

\paragraph{Key dimensions at fixed state size.} 
In our main experiments, we keep the first key dimension $d_k$ fixed at 128 and only vary the second key dimension from $E=1$ to $E=8$. Table~\ref{tab:ablations_400m} ablates this choice for Triadic GDN with $8\times$ the state. We vary the first key dimension $d_k$ against the second key dimension $E$ at a fixed joint key dimension $d_k \cdot E = 1024$, while keeping the value dimension and the number of heads fixed and using a single scalar forget gate per head. Since a smaller first key shrinks the total key and query projections, we match parameters by increasing the MLP width. As the two key dimensions become more similar, perplexity degrades slightly. Still, we believe that factorizations that require smaller total key and query projections could be relevant, in particular for mixture-of-experts architectures.

\paragraph{Second-key activation function.} 
We further ablate the activation function of the second key and query after the short convolution. Table~\ref{tab:ablations_400m} compares four activation functions for Triadic GDN with $8\times$ the state. Non-negative activation functions (softplus, sigmoid) consistently outperform activation functions that can take negative values (SiLU, no activation) in perplexity. One explanation is that signed entries let a token read and write to different slices with opposite signs, so their contributions can cancel out when the second query aggregates the slices.
\begin{table}[t]
\centering
\vspace{-2mm}
\scriptsize
\setlength{\tabcolsep}{3.2pt}
\renewcommand{\arraystretch}{1.08}
\caption{Ablations of Triadic Gated DeltaNet at 400M parameters. Top: first key dimension $d_k$ versus second key dimension $E$ at fixed joint key dimension $d_k \cdot E= 1024$, using a single forget gate per head. Bottom: activation function applied to the second key and query for $E=8$.}
\vspace{-1mm}
\label{tab:ablations_400m}
\resizebox{\linewidth}{!}{
\begin{tabular}{l|c|c|cc|ccc|c}
\toprule
\textbf{Model}
& \textbf{State}
& \textbf{Params}
& \textbf{0-shot $\uparrow$}
& \textbf{Wiki. $\downarrow$}
& \textbf{PG19 $\leq$4k $\downarrow$}
& \textbf{PG19 4k--16k $\downarrow$}
& \textbf{PG19 16k--64k $\downarrow$}
& \textbf{Recall $\uparrow$} \\
\midrule

Triadic GDN (scalar gate), $d_k = 128$, $E = 8$
& 50.3
& 384.0
& \textbf{53.7} & \textbf{10.80}
& \textbf{14.77} & \textbf{14.01} & \textbf{13.70}
& \textbf{33.4} \\

\quad $d_k = 64$, $E = 16$
& 50.3
& 380.8
& 52.9 & 10.84
& 14.79 & 14.04 & 13.73
& 32.5 \\

\quad $d_k = 32$, $E = 32$
& 50.3
& 383.9
& 53.3 & 10.90
& 14.82 & 14.08 & 13.78
& 32.5 \\

\quad $d_k = 32$, $E = 32$, tied keys
& 50.3
& 380.7
& 52.1 & 10.89
& 14.81 & 14.09 & 13.79
& 31.9 \\

\midrule

Triadic GDN ($E = 8$), Softplus activation
& 50.3
& 385.4
& 53.5 & 10.86
& \textbf{14.81} & \textbf{14.07} & \textbf{13.77}
& 32.2 \\

\quad Sigmoid activation
& 50.3
& 385.4
& \textbf{53.7} & \textbf{10.86}
& 14.82 & 14.07 & 13.77
& \textbf{33.3} \\

\quad SiLU activation
& 50.3
& 385.4
& 53.0 & 10.98
& 14.92 & 14.16 & 13.84
& 32.8 \\

\quad Linear (no activation)
& 50.3
& 385.4
& 53.0 & 11.06
& 15.05 & 14.29 & 13.98
& 32.9 \\

\bottomrule
\end{tabular}
}
\vspace{-2mm}
\end{table}

\section{Discussion and Limitations}
Our results indicate that state size is an important axis for improving linear RNNs, in particular  on recall-intensive tasks and at long context. Triadic linear attention provides a parameter-efficient 
way to increase it, similar to how (dyadic) linear attention increases the state size of traditional vector-valued RNNs. At both 400M and 1.3B scales, triadic linear attention consistently improves perplexity and recall, and it outperforms alternative methods to enlarge the state size. 

Nevertheless, the larger state slows down training by around 15\% for $E=4$ and by around 30\% for $E=8$, even with our optimized kernels. Additionally, performance on some recall-intensive tasks still trails that of Transformers, but only with a far larger state size at long context. Finally, we apply the triadic construction only to GDN and sGLA, which is only a small subset of the linear attention literature. Many other developments for matrix-valued states could be revisited for three-dimensional states, and the additional axis opens up possibilities for entirely new variants.

\vspace{-0mm}
\section{Related Work}
\paragraph{Linear attention.} Linear attention replaces the softmax over the attention logits with a kernel feature map. This allows the growing key-value cache to collapse into a fixed-size recurrent state. The resulting recurrence writes to a matrix state by an outer product of a key and a value and reads from it using the query \citep{katharopoulos2020transformers}. Up to the normalization factor omitted in modern variants, this update coincides with the fast weight programmer of \citet{schmidhuber1992learning}, as observed by \citet{schlag_linear_2021}. Early work on linear attention focused on better feature maps \citep{peng_2021_random, choromanski2021rethinking,arora2024simple}, a line of work directly relevant here, since our three-dimensional state can be interpreted as an ordinary matrix state under a feature map that expands the concatenated key $[k, k'] \in \mathbb{R}^{d+E}$ into the $d \cdot E$-dimensional key $k \otimes k'$. Subsequent work introduced forgetting to the recurrence, first through fixed decay \citep{sun2023retentive} and later through data-dependent gates \citep{yang2023gated, dao2024transformers, beckxlstm, qin_hgrn2_2024, peng2024eagle}. A complementary development adds the delta rule \citep{widrow_adaptive_1988} into linear attention, yielding DeltaNet \citep{schlag_linear_2021, yang2024parallelizing}. Combining this with data-dependent forgetting gives Gated DeltaNet \citep{yang2024gated}. More recent work generalizes the Gated DeltaNet recurrence to identity-plus-low-rank state transitions and adds more flexible parametrizations \citep{peng2025rwkv, hatamizadeh2026gateddeltanet2decouplingerase}.

\paragraph{Test-time training.}
Test-time training (TTT) interprets the in-context update of a recurrent state as learning under a self-supervised inner loss~\citep{sun2025learninglearntesttime}. For a linear inner model with squared reconstruction loss $\lVert \mathbf{S}_t k_t - v_t \rVert^2$, one gradient step per token recovers the delta rule~\citep{schlag_linear_2021}. DeltaProduct \citep{siems2025deltaproduct} instead performs multiple gradient descent steps per token, while MesaNet \citep{von2025mesanet} solves the regression to optimality at every step. These methods remain close to conventional linear attention and can be compared at matched state size. In contrast, many TTT variants, including TTT-MLP \citep{sun2025learninglearntesttime}, LaCT \citep{zhang2026test}, deep neural memories \citep{behrouz2025titans, behrouz2026atlas}, and TTT-E2E \citep{tandon2025endtoendtesttimetraininglong} use significantly larger states than conventional linear attention blocks. Our results show that increasing state size substantially improves linear attention. Since a gradient step can itself be interpreted as a linear-attention update~\citep{irie2022dual}, some of the gains of TTT variants may therefore be a result of their larger state capacity, particularly at long context lengths.

\vspace{-0mm}
\paragraph{Increasing state size.}
\citet{arora2024simple} identify a tradeoff between recurrent state size and recall, and we adopt their recall-focused benchmark suite. Dense linear-attention variants can increase state size through larger key or value dimensions, more heads, or multiple value heads per key head~\citep{Gu2023MambaLS, dao2024transformers, yang2025qwen3technicalreport}. These changes stay close to the original formulation and therefore serve as our main baselines in Section~\ref{sec:empiricalstudy}. A second line of work uses a fixed set of memory slots and routes over them \citep{peng-etal-2022-abc, zhang2024gatedslotattention, afzalbick2026raven}. Mixture-of-Memories extends this idea to multiple matrix-valued states with sparse routing \citep{du2026mom}. Related methods use product-key addressing \citep{lample2019productkeymemory, zhao2026fwpkm, cabannes2026sparsedeltamemoryscaling} to manage very large memories sparsely. These approaches generally have much higher state size compared to our construction, which comes with substantial computational overhead for memory movement. Triadic linear attention instead uses a single dense state and admits efficient fused kernels, which makes the increase in state size practical.
\vspace{-0mm}
\paragraph{Higher-order associative memories.}
The matrix state of linear attention stores associations as a sum of outer products and is thus a correlation matrix memory \citep{kohonen1972correlation}. In that literature, moving from pairwise to higher-order interactions is a standard route to higher capacity \citep{poggio1975optimal, chen1986high}. The number of storable patterns then scales as $d^{n-1}$, where $d$ is the number of memory units and $n$ is the interaction order \citep{baldi1987number}. This line of work was revived as dense associative memory \citep{krotov2016dense} and later related to attention \citep{ramsauer2021hopfield}. \citet{Smolensky1990Tensor} interpreted the same construction from a symbolic perspective, where a filler is bound to a role by an outer product and unbound by contraction. The closest prior work to ours is \citet{schlag2018thirdOrder}, who place a third-order state inside a recurrent network. As in our work, each write updates a third-order tensor with a tensor product, but their recurrent update also includes three operations specifically designed for graph traversal (write, move, backlink). Their work focuses on compositional reasoning on small-scale language tasks, while we use a similar construction to enlarge the state of a modern linear-attention layer. The resulting sequence mixer incorporates a multi-head architecture, short convolutions, nonlinearities, normalizations, data-dependent forgetting, and a learnable delta rule, while efficient kernels allow us to train stacks of these layers at scale for general language modeling.

\section{Conclusion}
\vspace{-1mm}
We introduced triadic linear attention, which updates a third-order tensor state with the outer product of three vectors and thereby increases the capacity of constant-time sequence mixers. Compared to other methods, this construction adds little parameter overhead and admits efficient kernels that exploit its structure. Across language modeling experiments at the 400M and 1.3B parameter scales, triadic Gated DeltaNet consistently lowers perplexity and improves recall, while custom kernels keep the training time within 1.3 times that of vanilla Gated DeltaNet for $E=8$. We hope that triadic linear attention is a starting point for linear attention variants with larger states and ultimately for pure linear attention models at the frontier.

\section*{Acknowledgments}
We thank Han Guo for valuable discussions and feedback. This study was supported by MIT-IBM Computing Research Lab and  the AI2050 program at Schmidt
Sciences (Grant G-25-67980).

\bibliography{iclr2027_conference}
\bibliographystyle{iclr2027_conference}
\newpage
\appendix

\section{Chunkwise Form and GPU kernels}
\label{app:kernels}
Section~\ref{subsec:efficient} derives the chunkwise form of triadic linear attention without forget gates and the delta rule.
Here we give the form with both, as Triadic GDN uses it, and describe the kernels that compute it.

\paragraph{Chunkwise form.}
Following \citet{yang2023gated, yang2024parallelizing}, we split the sequence into chunks of $C = 64$ positions indexed by $i$, write $\square^r = \square_{iC+r}$ for the $r$-th position of chunk $i$, and write $\mathbf{S}_{[i],e} \in \mathbb{R}^{d \times d}$ for slice $e$ of the state at the start of chunk $i$.
We drop the chunk index on all other quantities of the current chunk.
Its keys, queries and values are stacked into $\mathbf{K}, \mathbf{Q}, \mathbf{V} \in \mathbb{R}^{C \times d}$, its second keys and queries into $\mathbf{K}^{\prime}, \mathbf{Q}^{\prime} \in \mathbb{R}^{C \times E}$, whose columns $\mathbf{k}^{\prime}_e, \mathbf{q}^{\prime}_e \in \mathbb{R}^{C}$ belong to slice $e$, and its write strengths into $\boldsymbol{\beta} \in \mathbb{R}^{C}$.
The decay that slice $e$ accumulates over the first $r$ positions of the chunk is $\gamma^r_e = \prod_{j=1}^{r} \alpha^j_e$, and $\boldsymbol{\gamma}_e \in \mathbb{R}^{C}$ collects it.
As in Gated DeltaNet \citep{yang2024gated}, each slice has a decay matrix $\boldsymbol{\Gamma}_e \in \mathbb{R}^{C \times C}$ with $(\boldsymbol{\Gamma}_e)_{rs} = \gamma^r_e / \gamma^s_e$ for $r \ge s$ and $0$ otherwise.
The part of the joint key $\boldsymbol{\kappa}^r = \mathbf{k}^r \otimes \mathbf{k}^{\prime r}$ that addresses slice $e$ is $k^{\prime r}_e \mathbf{k}^r$, and each slice decays by a single scalar, so the decayed inner product of two joint keys still separates into a term of the keys and a term of the second keys,
\begin{equation}
  \sum_{e=1}^{E} \frac{\gamma^r_e}{\gamma^s_e} \big( k^{\prime r}_e \mathbf{k}^r \big)^{\top} \big( k^{\prime s}_e \mathbf{k}^s \big)
  = \big( \mathbf{k}^{r\top} \mathbf{k}^s \big)\, R_{rs},
  \qquad
  \mathbf{R} = \sum_{e=1}^{E} \big( \mathbf{k}^{\prime}_e \mathbf{k}^{\prime\top}_e \big) \odot \boldsymbol{\Gamma}_e,
  \qquad r \ge s.
  \label{eq:joint-gram}
\end{equation}
The UT transform of the chunkwise delta rule is therefore formed from products of the $d$-dimensional keys and the $C^2 E$ terms of $\mathbf{R}$, rather than from the $d \cdot E$-dimensional joint keys,
\begin{equation}
  \mathbf{T} = \Big( \mathbf{I} + \operatorname{tril}\big( \operatorname{diag}(\boldsymbol{\beta})\,
  (\mathbf{K}\mathbf{K}^{\top} \odot \mathbf{R}),\, -1 \big) \Big)^{-1} \operatorname{diag}(\boldsymbol{\beta}),
  \label{eq:joint-solve}
\end{equation}
and it is the same $C \times C$ matrix for all $E$ slices.
With $\mathbf{R}^{\prime} = \sum_e (\mathbf{q}^{\prime}_e \mathbf{k}^{\prime\top}_e) \odot \boldsymbol{\Gamma}_e$ for the query side, one chunk computes
\begin{equation}
\begin{aligned}
  \mathbf{U} &= \mathbf{T} \Big( \mathbf{V} - \textstyle\sum_{e} \operatorname{diag}(\mathbf{k}^{\prime}_e \odot \boldsymbol{\gamma}_e)\, \mathbf{K} \mathbf{S}_{[i],e} \Big), \\
  \mathbf{O} &= \textstyle\sum_{e} \operatorname{diag}(\mathbf{q}^{\prime}_e \odot \boldsymbol{\gamma}_e)\, \mathbf{Q} \mathbf{S}_{[i],e}
     + \big( \mathbf{Q}\mathbf{K}^{\top} \odot \mathbf{R}^{\prime} \big)\, \mathbf{U}, \\
  \mathbf{S}_{[i+1],e} &= \gamma^C_e\, \mathbf{S}_{[i],e} + \mathbf{K}^{\top} \operatorname{diag}\big( \mathbf{k}^{\prime}_e \odot \gamma^C_e / \boldsymbol{\gamma}_e \big)\, \mathbf{U},
\end{aligned}
\label{eq:joint-chunk}
\end{equation}
where the division acts elementwise and $\mathbf{U}$ holds the chunk's delta-corrected values.
Without forget gates, every $\boldsymbol{\Gamma}_e$ is the causal mask, and $\mathbf{R}^{\prime}$ reduces to $\operatorname{tril}(\mathbf{Q}^{\prime}\mathbf{K}^{\prime\top})$ of Section~\ref{subsec:efficient}; for $E = 1$ and $\mathbf{k}^{\prime} = \mathbf{q}^{\prime} = \mathbf{1}$, $\mathbf{R} = \mathbf{R}^{\prime} = \boldsymbol{\Gamma}$ and Equation~\ref{eq:joint-chunk} is the chunkwise form of Gated DeltaNet.
Unlike DeltaNet, which precomputes $\mathbf{W} = \mathbf{T}\mathbf{K}$ and $\mathbf{T}\mathbf{V}$ before the recurrence \citep{yang2024parallelizing}, we apply $\mathbf{T}$ to the residual, since with per-slice decay the analogue of $\mathbf{W}$ would be one matrix $\mathbf{T}\operatorname{diag}(\mathbf{k}^{\prime}_e \odot \boldsymbol{\gamma}_e)\mathbf{K}$ per slice.

\paragraph{Numerical stability.}
The per-slice gates make the decay vary along the joint key, which in gated linear attention requires a secondary level of chunking \citep{yang2023gated}.
Here each slice decays by a single scalar, so the decay enters only the $C^2 E$ terms of $\mathbf{R}$ and $\mathbf{R}^{\prime}$ and the per-slice scalings in Equation~\ref{eq:joint-chunk}, and the kernels evaluate each decay factor, such as $\gamma^r_e / \gamma^s_e$, directly as the exponential of a difference of accumulated log gates.
Every such difference is non-positive, because the log decay only decreases within a chunk and $\boldsymbol{\Gamma}_e$ is used only for $r \ge s$.
The computation therefore cannot overflow at any decay rate, and the gates need no clamp.
This rules out a cheaper construction of $\mathbf{R}$.
Writing each factor as $(\gamma^r_e / m_e)(m_e / \gamma^s_e)$ around a reference value $m_e$ in the middle of the chunk turns $\mathbf{R}$ into the lower triangle of a product of two $C \times E$ matrices and reduces the number of exponentials from $C^2 E$ to $2CE$, but gives the factors positive exponents, which overflow in FP32 above about 88.7.
With the exponents clamped to $[-88, 88]$, once a slice decays by more than a factor of $e^{176}$ within one chunk, the clamp distorts the near-diagonal entries of $\mathbf{R}$.
In an FP64 simulation of one chunk that emulates only this clamp, the relative error of the chunk's output is $0.05$ at a within-chunk decay of $e^{-200}$ and $1.5$ at $e^{-600}$.
We therefore form $\mathbf{R}$ and $\mathbf{R}^{\prime}$ element by element.

\paragraph{Input projections.}
The second keys and queries share the projection and the short convolution of the queries, keys and values: one GEMM produces $\mathbf{q}, \mathbf{k}, \mathbf{v}, \mathbf{k}^{\prime}$ and $\mathbf{q}^{\prime}$, and one causal depthwise convolution processes all of them, with SiLU on the channels of $\mathbf{q}$, $\mathbf{k}$ and $\mathbf{v}$ only.
A triadic layer therefore runs the same two projection GEMMs and one convolution as a GDN layer, with wider projections.

\paragraph{Kernels.}
We implement the chunkwise form for Hopper GPUs in the CuTe DSL of CUTLASS \citep{Thakkar_CUTLASS_2023}.
The forward pass accumulates the log decays, forms $\mathbf{R}$ and $\mathbf{R}^{\prime}$, computes the inverse in $\mathbf{T}$ with a blockwise triangular solve accumulated in FP32, and then runs one recurrent kernel over the chunks of each sequence, with one thread block per document, head and value block of 32 columns, or 16 when a step holds too few documents to fill the GPU.
This kernel is warp-specialized.
At $E = 8$, two warpgroups hold four state slices each and compute their products with asynchronous tensor-core instructions, a third warpgroup loads each chunk's inputs through the Tensor Memory Accelerator into double-buffered shared memory and forms $\mathbf{U}$, and a fourth forms the output.
For training, the forward pass also stores the state at the start of every chunk in BF16, so that the backward pass does not recompute it.
The backward pass runs a short reverse recurrence over the chunks that stores the gradient of each state slice, after which all remaining gradients are computed for every chunk in parallel.
Every reduction follows a fixed order without floating-point atomics, so the gradients are bitwise reproducible.
Documents packed into one training sequence each run their own recurrence from a zero state, directly on the packed tensors.
We verify the output and the gradients of all seven inputs against FP64 references at the 1.3B training shape, with a relative $\ell_2$ error below 1\%.

\section{Analysis}
\label{app:analysis}

\begin{figure}[t]
    \centering
    \includegraphics[width=\linewidth]{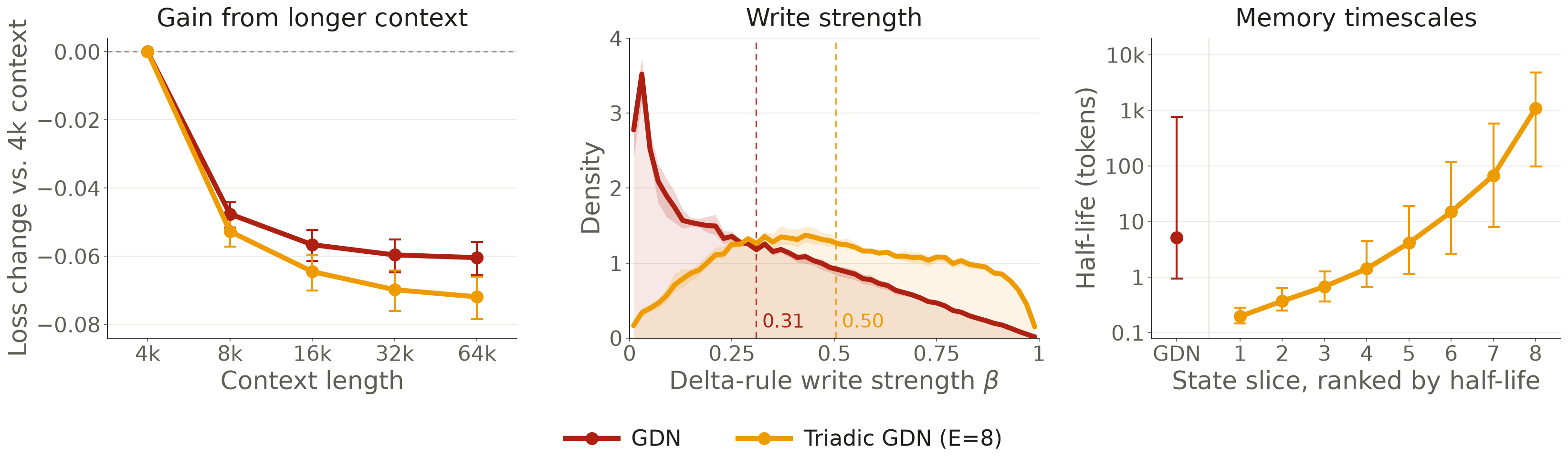}
    \caption{How Triadic GDN ($E{=}8$) uses the larger state compared to GDN. Left: change in loss on the final 4k tokens of a book when increasing the preceding context from 4k to 64k. Middle: distribution of delta strengths $\beta$. Dashed lines mark the means. Right: half-life of each state slice, ranked within its head.}
    \label{fig:analysis}
\end{figure}

\paragraph{Long-context utilization.}
A larger recurrent state is useful when it enables the model to preserve information from long ago. We test this directly on our trained long-context checkpoints of GDN and Triadic GDN ($E{=}8$) across three seeds on held-out PG19 books containing at least 64k tokens. Per book, we fix the prediction target to the last 4k tokens and vary the context the state reads before predicting them. As shown in Figure~\ref{fig:analysis} (left), Triadic GDN gains consistently more from added context than vanilla Gated DeltaNet. In addition, we find that GDN begins to saturate by 16k context tokens, while Triadic GDN continues improving through 32k.

\paragraph{Memory update dynamics.} 
We next examine how enlarging the state size changes how the model works with its memory. When comparing Triadic GDN ($E{=}8$) against GDN, we find that Triadic GDN learns substantially larger delta-rule write strengths $\beta$ than GDN, with the per-token $\beta$ distribution also becoming more broadly distributed. Averaged across all evaluated tokens, heads, and layers, the mean increases from 0.31 to 0.50 (Figure~\ref{fig:analysis}, middle). We also find that the additional state slices develop a broad range of memory timescales. To quantify this, we measure each slice's half-life, or the number of tokens over which the forget gates reduce an existing state contribution by half. Ranking the eight slices within each head by this half-life (Figure~\ref{fig:analysis}, right), the median half-life ranges from 0.20 tokens for the shortest-lived slice to 1080 tokens for the longest-lived slice. In comparison, GDN only has a single forget gate per head, with a median half-life of 5.1 tokens. Together, these suggest that the additional state axis permits the model to make stronger memory updates while keeping information over a wider range of timescales.

\section{MQAR Setup}
\label{app:mqarsetup}

Each example of $N$ key-value pairs yields a sequence with $2N$ tokens. The first $N$ positions contain the $N$ key-value pairs in random order. The next $N$ positions contain the $N$ queries (all keys corresponding to a key-value pair in the first $N$ positions) and the model has to predict the corresponding value. To construct a sequence, keys and values are drawn without replacement from two distinct sets of 8k tokens. Each key-value pair is represented as a single token by concatenating the embedding of the key with the embedding of the value. A query is represented by concatenating the embedding of the key with zeros. Queries are read-only, meaning they never update the recurrent state, which is the linear-attention equivalent of masking out other queries in the attention mask of a transformer. Every query thus reads the same accumulated state.
\begin{algorithm}[t]
\caption{Block with vanilla multi-head triadic linear attention ($H$ heads, key/value dimension $D$, second-key dimension $E$)}
\label{alg:mixer}
\begin{algorithmic}[1]
\algrenewcommand{\algorithmiccomment}[1]{\textcolor{gray}{$\triangleright$ \textit{#1}}}
\Statex \algorithmiccomment{Vanilla triadic linear attention}
\State $u \gets \mathrm{RMSNorm}(x_t)$
\For{$h = 1, \dots, H$}
  \State $q, k, v \gets W_q^h u,\; W_k^h u,\; W_v^h u \in \mathbb{R}^{D}$
  \State $q', k' \gets W_{q'}^h u,\; W_{k'}^h u \in \mathbb{R}^{E}$
  \State $S_h \gets S_h + k \otimes k' \otimes v$
  \State $o_h \gets S_h \times_1 q \times_2 q'$
\EndFor
\State $x_t \gets x_t + W_o\, [\mathrm{RMSNorm}(o_1); \dots; \mathrm{RMSNorm}(o_H)]$
\Statex
\Statex \algorithmiccomment{SwiGLU MLP}
\State $u \gets \mathrm{RMSNorm}(x_t)$
\State $x_t \gets x_t + W_{\mathrm{down}}\big(\mathrm{SiLU}(W_{\mathrm{gate}} u) \odot W_{\mathrm{up}} u\big)$
\end{algorithmic}
\end{algorithm}

Our model architecture consists of a vanilla variant of triadic linear attention interleaved with SwiGLU MLPs \citep{shazeer2020glu}. We provide pseudocode of the resulting block in Algorithm~\ref{alg:mixer}.
The sequence mixer is a pure outer-product update of a third-order tensor, without any forgetting, convolutions, delta rule or non-linearities (beyond RMSNorm). We choose $d_{\text{model}} = 128$, $H = 4$ heads, key/value dimension $D = 16$, SwiGLU width $384$, and a vocabulary size of $16$k ($8$k key tokens, $8$k value tokens). The second-key dimension $E$ varies over $\{1, 2, 4, 8, 16\}$, which induces a per-head state tensor of shape $16 \times E \times 16$. $E = 1$ recovers ordinary (two-dimensional) linear attention. All models range from $3.51$M to $3.54$M parameters, where the majority ($3.15$M) sit in the embedding table and output head.

We train for $50$k steps with a batch size of $250$k tokens per step, yielding $12.5$ billion tokens in total. Every step we sample $\lfloor 250\mathrm{k} / 2N \rfloor$ fresh sequences of length $2N$. We use AdamW with learning rate $10^{-3}$, $\beta = (0.9, 0.999)$, weight decay $0.1$, cosine decay to $0$, no gradient clipping, and bfloat16. We sweep $N \in \{32, 64, \dots, 4096\}$ and $E \in \{1, 2, 4, 8, 16\}$ for $25$ seeds per cell and present the mean test accuracy over all seeds, measured on a held-out set of $3$k sequences, in Figure~\ref{fig:mqar}.

\section{Language Modeling Setup}
\label{app:lmsetup}
\paragraph{Model architecture.}
All models consist of 24 pre-norm blocks with RMSNorm, a sequence mixer and a SwiGLU MLP. Algorithm~\ref{alg:lm-mixer} gives the GDN and sGLA blocks. Since the second query and key are l2-normalized, at $E=1$ they are both 1 and thus the triadic variant reduces to the base mixer. The Transformer \citep{vaswani2017attention} uses grouped-query attention \citep{shazeer2019mqa, ainslie2023gqa} with 8 query heads per key-value head, QK-norm \citep{pmlr-v202-dehghani23a} and RoPE \citep{su2021roformer} with base theta 10k during pretraining and 2M for extension. We found that both QK-norm and the higher base frequency at extension (compared to 10k, 500k, 1M) strengthen the Transformer baseline. All models use the Llama-2 tokenizer \citep{touvron2023llama2openfoundation} with 32k vocabulary size.

\paragraph{Evaluation.}
 We evaluate the final checkpoint of the long-context extension. PG19 perplexity is measured on around 1700 held-out books on windows of 64k tokens, which yields around 150M tokens in total. For WikiText, we report per-token perplexity on the WikiText-2 test set, with each document passed into the model in its entirety. The zero-shot score we report is the average accuracy over LAMBADA \citep{paperno_lambada_2016}, HellaSwag \citep{zellers2019hellaswag}, PIQA \citep{bisk2020piqa}, ARC-Easy \citep{arc-ce}, ARC-Challenge \citep{arc-ce}, WinoGrande \citep{sakaguchi2021winogrande}, OpenBookQA \citep{mihaylov2018openbookqa}, SciQ \citep{welbl2017sciq}, BoolQ \citep{clark2019boolq} and COPA \citep{roemmele2011copa}, obtained through the lm-eval harness \citep{eval-harness}. We use length-normalized accuracy for HellaSwag, ARC-Challenge and OpenBookQA, and plain accuracy for the other tasks. For needle-in-a-haystack retrieval, we use the eight NIAH tasks of RULER \citep{hsieh2024ruler} at context lengths from 1k to 64k tokens and report the average over all tasks and lengths in Table~\ref{tab:hybrid_400m}. Figure~\ref{fig:niah_13b} shows four representative tasks, where the model retrieves a single needle hidden in essays (niah\_single\_2), one of four needles with different keys (niah\_multikey\_1), all four of four needles (niah\_multiquery), or four values stored under the same key (niah\_multivalue). For recall, we use the six tasks of \citet{arora2024simple}, namely DROP, NQ, TriviaQA, FDA, SQuAD and SWDE, and score an answer as correct if the generation contains it.

\begin{algorithm}[t]
\caption{Block with triadic GDN / sGLA as used in our language modeling experiments ($H$ heads, key/value dimension $D$, second-key dimension $E$, state $S_h \in \mathbb{R}^{D \times E \times D}$ per head)}
\label{alg:lm-mixer}
\begin{algorithmic}[1]
\algrenewcommand{\algorithmiccomment}[1]{\textcolor{gray}{$\triangleright$ \textit{#1}}}
\Statex \algorithmiccomment{Triadic linear attention with forgetting and (for GDN) the delta rule}
\State $u \gets \mathrm{RMSNorm}(x_t)$
\For{$h = 1, \dots, H$}
  \State $q, k, v \gets \mathrm{SiLU}(\mathrm{Conv}(W_q^h u)),\; \mathrm{SiLU}(\mathrm{Conv}(W_k^h u)),\; \mathrm{SiLU}(\mathrm{Conv}(W_v^h u)) \in \mathbb{R}^{D}$
  \State $q', k' \gets \mathrm{softplus}(\mathrm{Conv}(W_{q'}^h u)),\; \mathrm{softplus}(\mathrm{Conv}(W_{k'}^h u)) \in \mathbb{R}^{E}$
  \State $q, k, q', k' \gets q/\|q\|,\; k/\|k\|,\; q'/\|q'\|,\; k'/\|k'\|$
  \State $\alpha_e \gets \exp\!\big(-\exp(A^h_e)\,\mathrm{softplus}(W_{\alpha}^{h,e} u + b^h_e)\big)$ for $e = 1, \dots, E$
  \State $\beta \gets \sigma(W_\beta^h u)$
  \State $S_h \gets S_h \times_2 \mathrm{diag}(\alpha)$
  \If{GDN}
    \State $S_h \gets S_h + \beta\, k \otimes k' \otimes \big(v - S_h \times_1 k \times_2 k'\big)$
  \ElsIf{sGLA}
    \State $S_h \gets S_h + k \otimes \beta k' \otimes v$
  \EndIf
  \State $o_h \gets S_h \times_1 q \times_2 q'$
  \State $o_h \gets \mathrm{RMSNorm}(o_h) \odot \sigma\!\big(W_{g,2}^h W_{g,1} u + b_g^h\big)$
\EndFor
\State $x_t \gets x_t + W_o\, [o_1; \dots; o_H]$
\Statex
\Statex \algorithmiccomment{SwiGLU MLP}
\State $u \gets \mathrm{RMSNorm}(x_t)$
\State $x_t \gets x_t + W_{\mathrm{down}}\big(\mathrm{SiLU}(W_{\mathrm{gate}} u) \odot W_{\mathrm{up}} u\big)$
\end{algorithmic}
\end{algorithm}

\end{document}